\documentclass[lettersize,journal]{IEEEtran}
\usepackage{amsmath,amsfonts}
\usepackage{array}

\usepackage{textcomp}
\usepackage{stfloats}
\usepackage{url}
\usepackage{verbatim}
\usepackage{graphicx}
\usepackage{cite}
\usepackage[T1]{fontenc}
\usepackage{tabularx}
\usepackage{graphicx}

\usepackage{subcaption}
\usepackage{xcolor}
\usepackage[para]{footmisc}
\usepackage{tabularx,array}
\usepackage{multirow}
\usepackage{booktabs}
\usepackage[linesnumbered,ruled,vlined]{algorithm2e}
\SetKwInput{KwInput}{Input}                % Set the Input
\SetKwInput{KwOutput}{Output}  
\SetKwInput{KwFunction}{Function}  
\usepackage{textcomp}
\usepackage{linguex}
\usepackage[linguistics,edges]{forest}

\usepackage{metalogo}
\usepackage{url}
\usepackage{csquotes} % for double quotation
\usepackage{arydshln} % dash line used in table
\usepackage{color, soul}
\usepackage{amssymb}
\usepackage{makecell}
\usepackage{etoolbox}
\usepackage{threeparttable}
\usepackage{pifont}% http://ctan.org/pkg/pifont

\usepackage{float}
\usepackage[normalem]{ulem}

\usepackage{balance}

\begin{document}

\title{GFCL: A GRU-based Federated Continual Learning Framework against Data Poisoning Attacks in IoV}

\author{Anum~Talpur,~\IEEEmembership{Member,~IEEE,}
	and~Mohan~Gurusamy,~\IEEEmembership{Senior~Member,~IEEE}% <-this % stops a space
	\thanks{A. Talpur and M. Gurusamy are with the Department of Electrical and Computer Engineering, National University of Singapore, Singapore (email: anum.talpur@u.nus.edu; gmohan@nus.edu.sg).}}% <-this % stops a space

% make the title area
\maketitle

\begin{abstract}
Integration of machine learning (ML) in 5G-based Internet of Vehicles (IoV) networks has enabled intelligent transportation and smart traffic management. Nonetheless, the security against adversarial poisoning attacks is also increasingly becoming a challenging task. Specifically, Deep Reinforcement Learning (DRL) is one of the widely used ML designs in IoV applications. The standard ML security techniques are not effective in DRL where the algorithm learns to solve sequential decision-making through continuous interaction with the environment, and the environment is time-varying, dynamic, and mobile. In this paper, we propose a Gated Recurrent Unit (GRU)-based federated continual learning (GFCL) anomaly detection framework against Sybil-based data poisoning attacks in IoV. The objective is to present a lightweight and scalable framework that learns and detects the illegitimate behavior without having a-priori training dataset consisting of attack samples. We use GRU to predict a future data sequence to analyze and detect illegitimate behavior from vehicles in a federated learning-based distributed manner. We investigate the performance of our framework using real-world vehicle mobility traces. The results demonstrate the effectiveness of our proposed solution in terms of different performance metrics.

\end{abstract}


\section{Introduction}
With the adoption of modern networking and 5G technologies, the internet of vehicles (IoV) brings in enormous benefits to the automative industry \cite{IOVML2}. The concept of cellular vehicle-to-everything (C-V2X) was first standardized in the 3rd generation partnership project (3GPP) release~14. Today, in the era of 5G, we witness promising developments in IoV. It contributes to broader connectivity, fast responses, wider interoperability, high reliability, and smart traffic management. To achieve high performance, IoV requires intelligent decision-making capabilities. Thus, integration of machine learning (ML) and artificial intelligence (AI) is a promising approach for IoV applications. \par 
Innovations and intelligence always bring advantages and limitations. The adversarial attacks on ML are one such limitation that is becoming an active area of research as ML grows to play an important role in vehicle automation. While ML provides transportation intelligence and improved safety in IoV, the security of the ML model itself is not guaranteed. Based on a recent survey, Microsoft shares that 90\% of businesses don't have enough security for their ML-operated systems \cite{microsoft}. Similarly, the adversarial attacks, such as data poisoning attacks, on ML in the context of vehicular industry are not much explored in the literature \cite{anumsurvey}. Such attacks in vehicular applications can be a danger to human lives leading to dire consequences unless the possible vulnerabilities are not addressed. \par
\begin{figure}[htbp]
	\centering
	\includegraphics[width=3.2in, height=1.7in]{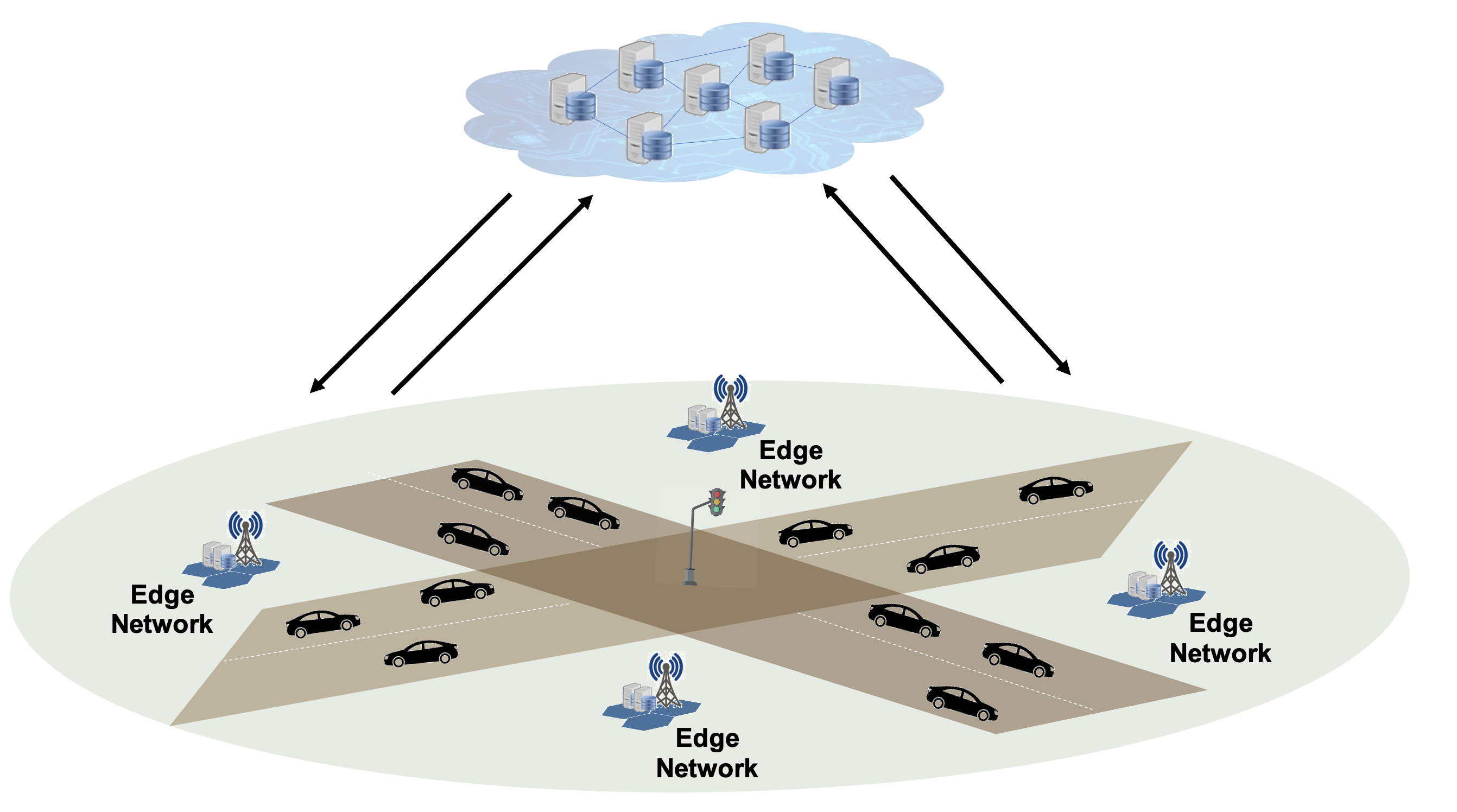}
	\caption{The architecture of IoV Network}
	\label{fig:Archi}
\end{figure}

Driven by the time-varying and mobile nature of vehicles, DRL has become a promising ML technique for autonomous driving, task offloading, task scheduling, service placement, security and many other applications of vehicular networks \cite{RLILdriving,deepsurvey}. In this paper, we focus on data poisoning attacks on DRL. Fig. \ref{fig:Archi} shows a framework of the IoV network where vehicles communicate with the edge network to avail services from DRL-based intelligent control methods. The DRL framework consists of an agent and environment, where the environment generates a state (input data) and reward (feedback), and an agent takes an action based on its policy. Here, an agent is a deep learning model which repeatedly interacts with an environment (in the form of state and reward collection) and learns a policy. The continual interaction of the DRL framework with the environment (i.e. vehicles), exposes it to many security vulnerabilities. The security problems of DRL are different from other ML types because DRL learns to solve sequential decision-making through continuous feedback from the environment whereas traditional ML deals with the prediction problems. \par 
Several adversarial data poisoning attacks are common for vehicular applications and the impact of such attacks can increase with Sybils leading to denial of service due to overloading of the network with poisonous messages. The adversary type we study here performs data poisoning over the environment to interfere with the network's healthy functioning. The attack over an edge-based DRL framework is a serious threat where decisions are based on feedback/rewards collected by the DRL agent. In this paper, we develop an anomaly detection framework against Sybil-based data poisoning attacks in IoV. The objective of our work is to implement a life-long and lightweight anomaly detection framework for IoV applications where the environment is time-varying and dynamic. The goal is to propose a framework that learns and detects the illegitimate behavior without having a-priori training dataset consisting of attack samples. Several ML-based attack detection frameworks make specific assumptions about an adversary, prepare a supervised dataset and use a traditional classification model to train a network and detect the known pattern. Different from them, we propose a model which learns and detects an attack from the time-varying IoV environment as it progresses. \par 
With this, we present a gated recurrent unit (GRU)-based federated continual learning (GFCL) anomaly detection framework against data poisoning attacks. We make the following contributions in this paper. 
\begin{itemize}
	\item We use a GRU-based recurrent neural network (RNN) architecture in a sequence-to-sequence regression manner and predict a future data sequence to analyze and detect illegitimate behavior from vehicles. The attractive features of GRU include computational efficiency and low memory overhead which make our design lightweight and time-efficient in attack detection.
	\item We integrate a federated learning approach for data sharing in the IoV network where data is distributed. The idea is to enable the participation of more edge nodes and maximize the knowledge for efficient attack detection design. 
	\item We run our model in a life-long learning (or continual learning) manner with the goal of extending the acquired stream of knowledge and using it for future misbehavior detection.
	\item We carry out performance evaluation on real-world IoV traces to verify the effectiveness of our proposed GFCL framework.
\end{itemize}
The rest of the paper is organized as follows. Section \ref{Sec:relatedwork} provides an overview of the related work in the literature. Section \ref{Sec:background} briefly describes the background on DRL, its security vulnerabilities, attack model, and solution approach. Section \ref{Sec:GRUsection} presents the architecture and GRU-based anomaly detection framework. Section \ref{Sec:FL} explains the integration of federated learning into our proposed GRU-based framework. Section \ref{Sec:performance-evaluation} presents the performance study and discusses the results. Section \ref{Sec:conclusion} makes concluding remarks.

\section{Related Work}
\label{Sec:relatedwork}
The continuous interaction of DRL with the environment makes it more vulnerable than other ML techniques. Data poisoning attacks like label flipping, backdoor attack, and model poisoning attack are very common adversarial attacks on DRL and are explored in the literature for IoV applications \cite{anumsurvey}. In a recent work, the authors provide a comprehensive survey on various adversarial data poisoning attacks on DRL \cite{DRLattacksurvey}. The authors in this work briefly discuss the possible choices of attacks carried out by an adversary on a DRL model which includes an attack on the state, action, reward, and model. Among the various works in the literature, the works that focus on attacks over reward function is recent. To the best of our knowledge, no works consider the vehicular applications where the environment is dynamic and time-varying. In this context, we study the problem of Sybil-based data poisoning attacks over the reward function of the DRL framework. \par
The Sybil attack is well-studied in the literature. However, the existing solutions are not effective on time-varying vehicular networks. This is due to the lack of the availability of the historical behavior in anomaly detection schemes and difficulty in tracing high mobility vehicles \cite{Sybilsurvey}. Some researchers propose to use privacy-preserving schemes against Sybil attacks \cite{privacy1,privacy2}. Nevertheless, adversaries are using well-trained intelligent ML techniques to craft an attack and impose more serious threats against privacy mechanisms in mobile networks \cite{ghost}\cite{attack}. Despite being a critical threat, the study on anomaly detection mechanisms against Sybil-based data poisoning attacks in IoV applications have not been explored much in the literature. \par 
In a recent work \cite{datapoisoningIOV}, Chen et. al presents state-of-the-art on data poisoning attacks in IoV applications. In \cite{firstadversarial}, Huang et. al studied the vulnerability of RL and shows that even in black-box scenarios it is easy to confuse RL policies. Another easy-to-deploy poisoning attack over a deep learning-enabled IoV application is performed by leveraging particle swarm optimization in \cite{swarm}. On the other hand, DRL can be used to provide defense and improve the recovery ability of the IoV network from data poisoning attacks \cite{TopologyPoisoning}. However, the defense against poisoning of DRL itself is less explored and requires investigation in the area of vehicular applications where traffic is mobile and time-varying.  \par 
The existing defense mechanisms against such attacks are classified into data-based defense and model-based defense \cite{adversarialdefensesurvey,datapoisoningIOV}. The data-based defense mechanisms perform adversarial training where they input adversarial examples in a supervised manner into the data while performing training. First, such defenses are effective when dealing with image datasets. Second, this method is not applicable for applications where we do not use existing big data and instead train our network from the environment as it progresses. In addition, several authors \cite{moosavi,AdamPoisoning,CrowdPoisoning} show that no matter how many adversarial examples we add, there are always new and improved techniques that can poison the network. \par 
In a model-based defense, the modification of the network structure is performed to improve robustness against poisoning attacks. In this method, sub-network or multimodel architectures are used to detect anomalies. In our work, we study the problem of attacks over the environment, the use of network structure modification is useful and effective by using an additional sub-network to detect anomalies and poisoned values before feeding data into the DRL framework. Authors in \cite{metzendetecting} use a deep neural network-based sub-network along with DRL to train and perform a binary classification for detecting real and adversarial data inputs. It achieves a detection accuracy of 70\% and works for image inputs. Siwakorn in recent work in \cite{thesis} proposes to use multiple models to enhance the robustness against adversarial attacks. This work uses a family of five models to enhance detection accuracy. It considers image datasets and works well for white-box poisoning attacks. \par 
Recently, researchers propose to use generative adversarial networks (GAN) to generate adversarial samples and train a neural network in a supervised manner to use an additional ML classifier before RL for detecting anomalous inputs \cite{hu2021rl}. Such methods are only effective when datasets are available beforehand. We note that GANs can be used not only for defense but also for performing adversarial attacks over RL \cite{attack2}. Cui et. al in \cite{defenseNew} propose a decentralized scheme that uses blockchain-based verification and federated learning along with DRL-based resource scheduling scheme to protect the DRL from being poisoned. This scheme is proposed for latency-sensitive applications with a fixed number of users without considering the mobility of users and its time-varying properties. \par 
In our work, we use a multiple-model design where a GRU-based federated continual learning (GFCL) is deployed before the DRL framework to help detect poisoned feedback values before inputting them into the model. Here, the choice of federated learning and continual learning along with GRU is significant because of distributed, mobile and time-varying properties of the vehicular environment.

\section{Attack Model and Proposed Solution Approach}
\label{Sec:background}
In this section, we present the background on DRL and its security vulnerabilities. Then, we describe the attack type and it's implementation. Finally, we discuss our proposed solution approach.

\subsection{Background}
\subsubsection{Overview of DRL}
\begin{figure}[htbp]
	\centering
	\includegraphics[width=2.6in, height=1.6in]{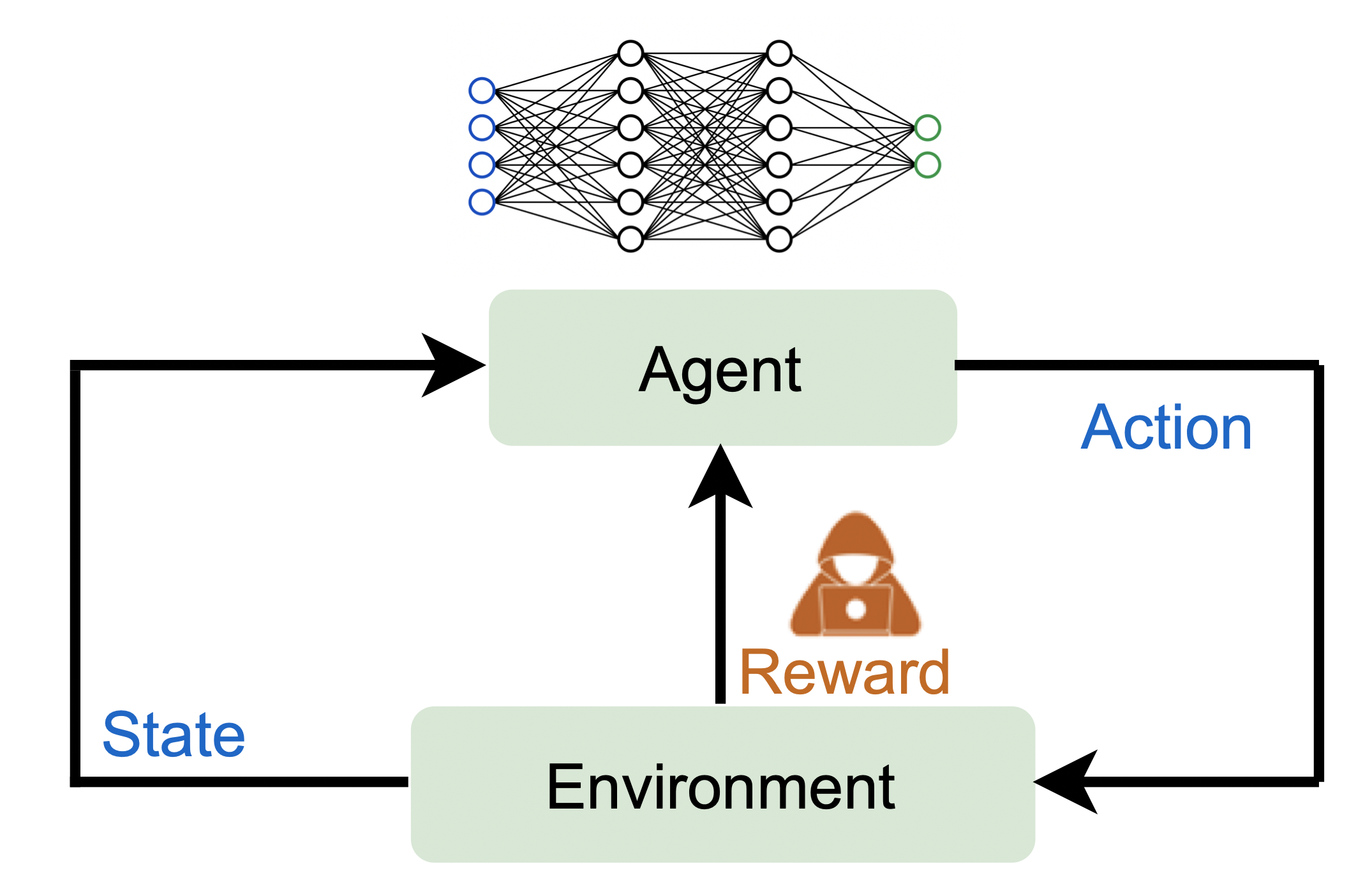}
	\caption{DRL Framework}
	\label{fig:drl}
\end{figure}

DRL uses a different plan of action to learn when compared to the traditional supervised and unsupervised learning frameworks. It is an objective-oriented learning framework. The conceptual diagram of DRL is shown in Fig. \ref{fig:drl}. It consists of an agent and environment, where the environment generates a state and reward, and the agent takes an action based on its policy. In DRL, the agent is a deep learning model which collects data (state) from an environment and learns a policy. Here, the policy is to maximize the network performance as it progresses in the form of positive future rewards. The agent uses a policy to take an action and in return, the environment gives feedback by rewarding the desired actions and punishing the undesired ones. \par 

\subsubsection{Vulnerabilities of DRL}
In DRL, where each time agent collects a reward to optimize its performance, the continual interaction of the agent's deep neural network framework with the environment exposes it to many security vulnerabilities. As a consequence, the attack on the environment and its generated rewards will result in undesirable behavior in the network. The assumption of a secure environment can be life-threatening that could result in accidents in DRL-based automation in vehicles. In our work, we investigate Sybil-based data poisoning attacks where the adversary has access to the environment, which can perform data poisoning over the rewards collected by the DRL agent to interfere with the network's healthy functioning.

\subsection{Attack and It's Implementation}
\label{Sec:AttackModel}
\subsubsection{Sybil-Based Data Poisoning Attack}
Several adversarial data poisoning attacks are common and explored in the literature for vehicular applications \cite{anumsurvey,DRLattacksurvey}. Different from the existing works, we use Sybil-based data poisoning attacks. A Sybil attack is an impersonation attack in which an illegitimate vehicle steals or borrows the identities of legitimate vehicles to execute an attack. Sybil attack is easy to launch in vehicular networks where the communication is wireless, broadcast, and dynamic. The adversary type we study here is a set of Sybil nodes that has access to the environment and performs data poisoning over the DRL reward values. 

\subsubsection{Implementation of Attack}
We assume a compromised vehicle/node that launches Sybil attacks and purposely stoles the identities of legitimate vehicles.  Several network monitoring tools exist that attackers can use to obtain and analyze the data. The stolen identities can easily be hidden by temporarily restricting the imitated vehicles from the network. We also assume that the adversary has access to all the information of the compromised vehicles which helps an attacker to pass the security checks of the network. Further, an adversary is able to reprogram the vehicle to perform data poisoning by sending forged information to the network. The forged information is the malicious reward values that can alter the optimal decision-making capability of the DRL framework. \par 
In our study, the reward values are delay values. Delay is one of the key performance metrics and is used by several research works as an objective parameter to optimize network performance. The goal of an adversary in this attack model is to manipulate the delay values and affect the model behavior where such values are aggregated to optimize network performance. The manipulation of delay values is interesting and noteworthy where we consider an attacker to be intelligent in poisoning data so that traditional rule-based methods or threshold-based methods cannot detect the attack. The rationale is the fake values are smaller delay values from the valid range of possible observed delay values in the given scenario. The choice of smaller values is random but from a valid range depending on the prior monitoring of traffic by an attacker because the goal is to choose poisonous delay values such that it is perceptually indistinguishable from true inputs values. The idea of replacing true delay values with smaller delay values will obscure the DRL framework, and it interprets that the network performance is good and vehicles can continue getting service from the same network settings. Thus the vehicles are receiving poor service and the delay is increasing which is undesirable. \par 
In our recent work in \cite{talpur2021adversarial}, we studied the above attack and analyzed its impact under different adversarial data poisoning attack scenarios over delay rewards for the DRL-based dynamic service placement application in IoV networks. We also examined the impact of the proportion of Sybil-attacked vehicles in the network and demonstrated that the performance is significantly affected by Sybil-based poisoning attacks when compared to adversary-free healthy network scenarios. Different from our previous work in \cite{talpur2021adversarial}, this work focuses on detecting such attacks and minimizing their impact on the network decision-making capabilities.

\subsection{Our Proposed Anomaly Detection Solution}
\label{Sec:ProblemandProposedSolution}
Having discussed the security vulnerabilities of DRL and the attack model, we present the attack detection framework in this section. We propose an effective solution to detect the vehicles under Sybil-based data poisoning attacks to exclude their participation in the aggregated rewards from the environment. We present a multifold design and develop a GRU-based federated continual learning (GFCL) anomaly detection framework with different objectives which are given below.
\begin{itemize}
	\item We use a GRU-based RNN architecture to design an attack detection framework at each edge node. The GRU-based learning model is used because of its ability to learn latent information. Another key reason behind using GRU in the proposed design is its ability to learn the relationship between data sequences and persist this information over time to help predict the future data sequence. This property of GRU helps us to use future data sequence and detect illegitimate behavior from vehicles. Also, the computational efficiency and low-overhead properties of GRU model make our design lightweight and time-efficient in attack detection. 
	\item Next, we integrate federated learning with GRU for efficient data sharing in the distributed IoV network. The use of FL maximizes the knowledge gained for attack detection by including more nodes in data collection process. FL integrated design also offers scalability as the training process is distributed among multiple edge nodes and executed in parallel. In addition, with the privacy-preserving property of the federated learning approach, data sharing is also secure.
	\item Finally, we run our model in a continual learning fashion. The idea is to extend the acquired stream of knowledge and use it for future misbehavior detection.
\end{itemize}

\section{GRU-based Anomaly Detection Framework}
\label{Sec:GRUsection}

\subsection{System Architecture}
\label{Sec:systemmodel}
\begin{figure}[htbp]
	\centering
	\includegraphics[width=3.1in, height=1.5in]{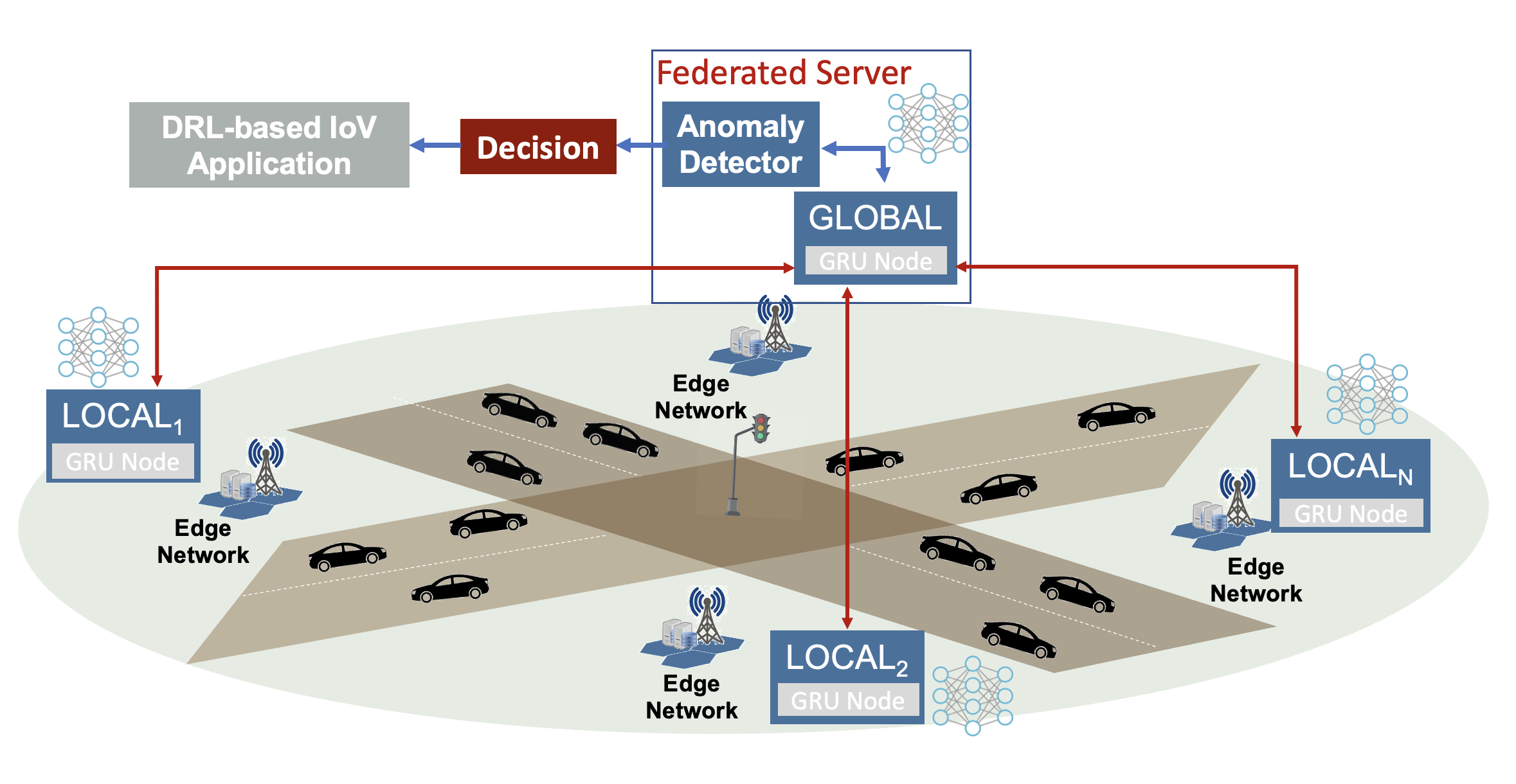}
	\caption{Our proposed anomaly detection framework}
	\label{fig:model}
\end{figure}

The architecture of our proposed IoV model is shown in Fig. \ref{fig:model}. It consists of an edge-enabled IoV environment. There is a data layer that consists of a city road environment with a real journey of taxis in the city of San Francisco. The vehicles are mobile, which are assumed to connect with edge nodes to avail different types of services or facilities from the IoV network. Vehicles can also connect to an edge network to get assistance in different tasks related to driving. We assume each vehicle $v$ is equipped with necessary sensors such as clock and GPS, which enable it to provide relevant information. The attacks are also assumed to take place at the data layer. The attack model is discussed in Section \ref{Sec:AttackModel}. \par
For the edge layer, we assume an IoV network environment with 5G coverage using evolved NodeB (eNB) stations. There are multiple eNBs equipped with edge servers that extend the capabilities (storage and compute) of the cloud and bring them closer to the end-user. Our GFCL anomaly detection framework (discussed in Section \ref{Sec:ProblemandProposedSolution}) is deployed at the edge layer. The edge nodes work in a cluster manner where there are multiple local nodes and one global node.  The local nodes deploy GRU-based design and send their learned model parameters to the global node. The global node is a federated server that has final decision-making capability where it uses its GFCL design with aggregated knowledge to give a decision on legitimate and illegitimate vehicle participation. Only those vehicles which are found legitimate will take part in the reward collection of the DRL-based IoV application. We do not consider using the cloud for deploying the global node here to avoid the additional delays which may add up because of frequent data transmission between long-distanced cloud and edge layers. Additionally, we assume adequate links between V2E (vehicle to edge) and E2E (edge to edge) nodes/servers are available to enable communication among them.

\subsection{GRU Node}
The structure of the GRU-based node consists of input data acquisition, pre-processing, and GRU model. 

\subsubsection{Data Acquisition and Pre-processing}
\label{Sec:preprocess}
Data acquisition is an important phase of an ML algorithm design. The vehicles in an IoV network generate massive data which captures the network states and characteristics that could be used by different DRL-based designs as input rewards to optimize the model performance. Delay is one of the potential performance indicators in IoV applications where Ultra-Reliable Low Latency Communications (URLLC)-based applications are gaining popularity. We choose to use delay as our input feature value. The selection of a single feature type is important in our application. The time duration of a vehicle connecting to the edge node is limited and keeps changing due to the vehicle's mobility. We use a single feature where multiple vehicles traveling along the same route and near to each other will generate the same range of delay values which are accumulated to generate enough data for efficient training of the deep model. Also, with the use of continual learning due to the time-varying nature of the IoV environment, the ML algorithm will be trained repeatedly to accommodate future changes. The use of multiple features will result in a massive amount of input and require a large amount of time to process data and generate results. Additionally, the choice of feature type is also important as irrelevant features may introduce noise and affect network performance. Therefore, our design limits the data acquisition phase to the collection of only delay values observed by vehicles while receiving a desired service type from the network. This also helps to reduce training complexity and achieve faster GRU-based predictions. \par 
Next, we pre-process our data to scale it into a definite range to improve the performance of our proposed anomaly detection framework. ML algorithms are known to perform better with a normalized set of input features. Note that because of normalization the proposed model is applicable even if the key feature is other than delay. Additionally, to prevent training data from diverging, it is important to remove large variations without changing the statistical properties of data. Therefore, we use Z-score data standardization to achieve scaling and prevent divergence. Using Z-score standardization, the data will be normalized to have standard distribution properties of zero mean and unit variance. This stage first involves the calculation of mean $\mu$ and variance $\sigma$ for the given set of input data. Then, each normalized feature value is obtained as,
\begin{equation}
\grave{x}=\frac{x-\mu}{\sigma}
\end{equation}
where, $x$ is the original input feature value and $\grave{x}$ is the respective standardized value. 

\subsubsection{GRU Model}
\label{Sec:GRU}
In this work, we use a variant of RNN, called as GRU neural network model, to predict the time-series data and perform anomaly detection. RNN is widely used in applications where the model needs to learn temporal information and process sequence data. Therefore, RNN and its variants are gaining promising applications in IoV networks. Vehicular networks are time-varying in nature and generate a long sequence of data with every new information showing relation with the previous one. This makes RNN one of the well-suited deep models for IoV networks. \par
GRU uses a \textit{gate} structure design to store or remove states from its memory cells. In our work, we use a typical GRU which contains two types of gates, hidden state, and current input. The two gate types are reset gate and update gate. These gates are two vectors that help keep useful information for a long time (using update gate) and remove the one which is irrelevant to the output prediction (using reset gate). Let $x$ be the current input and $h$ be the hidden input which holds the previous time slot value, then the update gate ($z_t$) parameters for time slot $t$ are calculated as,
%and  $y$ be output response,
\begin{equation}
z_t=\sigma(W^{z}x_t+U^{z}h_{t-1})
\end{equation}
where, $W^{z}$ and $U^{z}$ are the weights for current input and previous input, respectively. The sum of weighted inputs is then multiplied with $\sigma$ i.e. activation function to map the results between 0 and 1. The update gate also helps to eliminate the vanishing gradient problem. Next, the reset gate $r_t$ is calculated similar to the update gate,
\begin{equation}
r_t=\sigma(W^{r}x_t+U^{r}h_{t-1})
\end{equation}
where, $W^{z}$ and $U^{z}$ are the weights for current input and previous input of reset gate, respectively. As noticed, both gates have the same equation but a difference is observed in their usage. At first, the reset gate is used to create a current memory unit $h'_{t}$ as,
\begin{equation}
h'_{t}=tanh(Wx_t+r_t \odot Uh_{t-1})
\end{equation}
where $\odot$ is the Hadamard product. The $tanh$ is an activation function that maps the results between -1 and 1, to eliminate exploding gradient problem. In the end, the final memory unit at the current time slot is calculated,
\begin{equation}
h_{t}=(z_t \odot h_{t-1} + (1-z_t) \odot h'_{t})
\end{equation}
This final memory unit contains the previous and the most relevant information only. The suitable learning of $z_t $ and $1-z_t$ helps to adjust and decide on the weighted sum for previous information and current information to keep or remove. Due to its simple mathematical modeling, GRU is a computationally efficient and low memory overhead model which makes it suitable for a lightweight and time-efficient design.  \par 
The architecture of our proposed GRU-based neural network prediction model consists of an input layer, three GRU layers, and one fully-connected layer. First, we use pre-processed data as an input to the input layer. Let $X=\{x_1,x_2,...,x_n\}$ be the input sequence and $Y=\{y_1,y_2,...,y_n\}$ be the output sequence then we have to learn a mapping function where,
\begin{equation}
Y=f(X)
\end{equation}
The objective is to learn $f()$ so that we can predict the set of future $Y$ values when we have a set of input $X$ values. Our proposed design of sequence-to-sequence prediction use half-mean-squared-error (HMSE) as a loss function with the goal of minimizing the error between the actual and predicted value, 
\begin{equation}
loss=\frac{1}{2S}\sum_{i=1}^{S}\sum_{j=1}^{R}(y'_{ij}-y_{ij})
\label{Eq:loss}
\end{equation}
where $S$ is the input sequence length, $R$ is the output sequence length, $y$ is the predicted output and $y'$ is the targeted output. Then we use three-layer GRU architecture along with a dropout layer in-between to perform data prediction function. The function of the dropout layer is to neglect the selected percent of GRU cells data to escape from the local minima problem. Finally, there is a fully-connected layer followed by a classical regression layer to generate the output response. In our design, input is a batch $B$ of delay values experienced by different vehicles while accessing service from an edge node. Output is the predicted data sequence of delay values for the vehicles for the next $t'$ time slots. \par
We experiment our design with different layer sizes, input batch sizes, and learning rates. We then select the best values for the hyper-parameters which will be discussed in Section \ref{Sec:experimentalsetup}. The impact of different batch sizes is studied briefly in Section \ref{Sec:batchsize}.

\section{Integration of Federated Approach}
\label{Sec:FL}
Federated learning is a novel distributed machine learning framework that can be used for secure data sharing problems in IoV applications. FL integrated design offers scalability as the training process is distributed among multiple edge nodes and executed in parallel. In this work, we leverage a federated learning approach along with a GRU-based prediction framework for data sharing from vehicles where data is massively distributed. The key idea is to increase data participation from more edge nodes and maximize the knowledge for efficient attack detection design. \par
\begin{figure}[htbp]
	\centering
	\includegraphics[width=2.9in, height=1.1in]{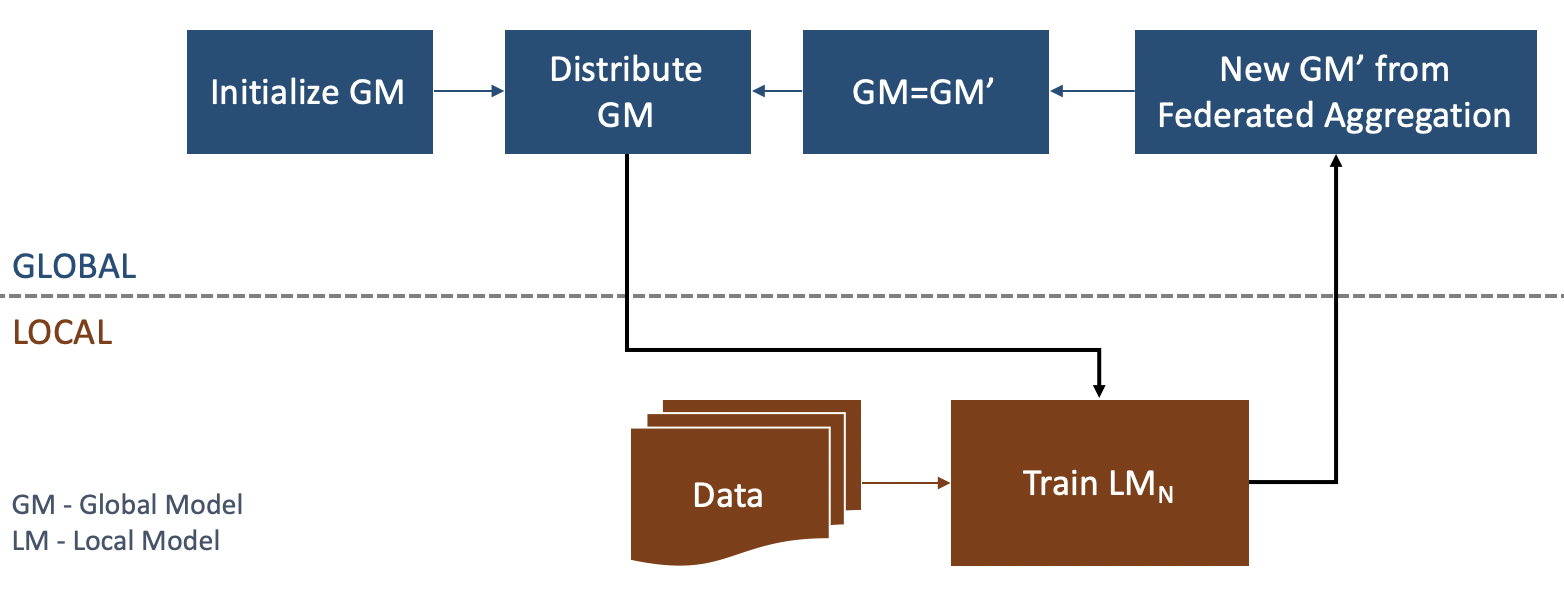}
	\caption{Continual federated learning}
	\label{fig:Fconcept}
\end{figure}

 
The conventional continual federated learning model offers learning updates between the server and clients, as shown in Fig. \ref{fig:Fconcept}. In our design, the client is a \textbf{\textit{local node}} at edge network which collects data from vehicles in its vicinity. The server is a \textbf{\textit{global node}} or a global server which builds the global model (GM) from the local models (LM) received from the local nodes. The acquired knowledge is used to optimize the network performance by minimizing HMSE between targeted and predicted delay values. \par
\begin{figure}[htbp]
	\centering
	\includegraphics[width=3in, height=1.6in]{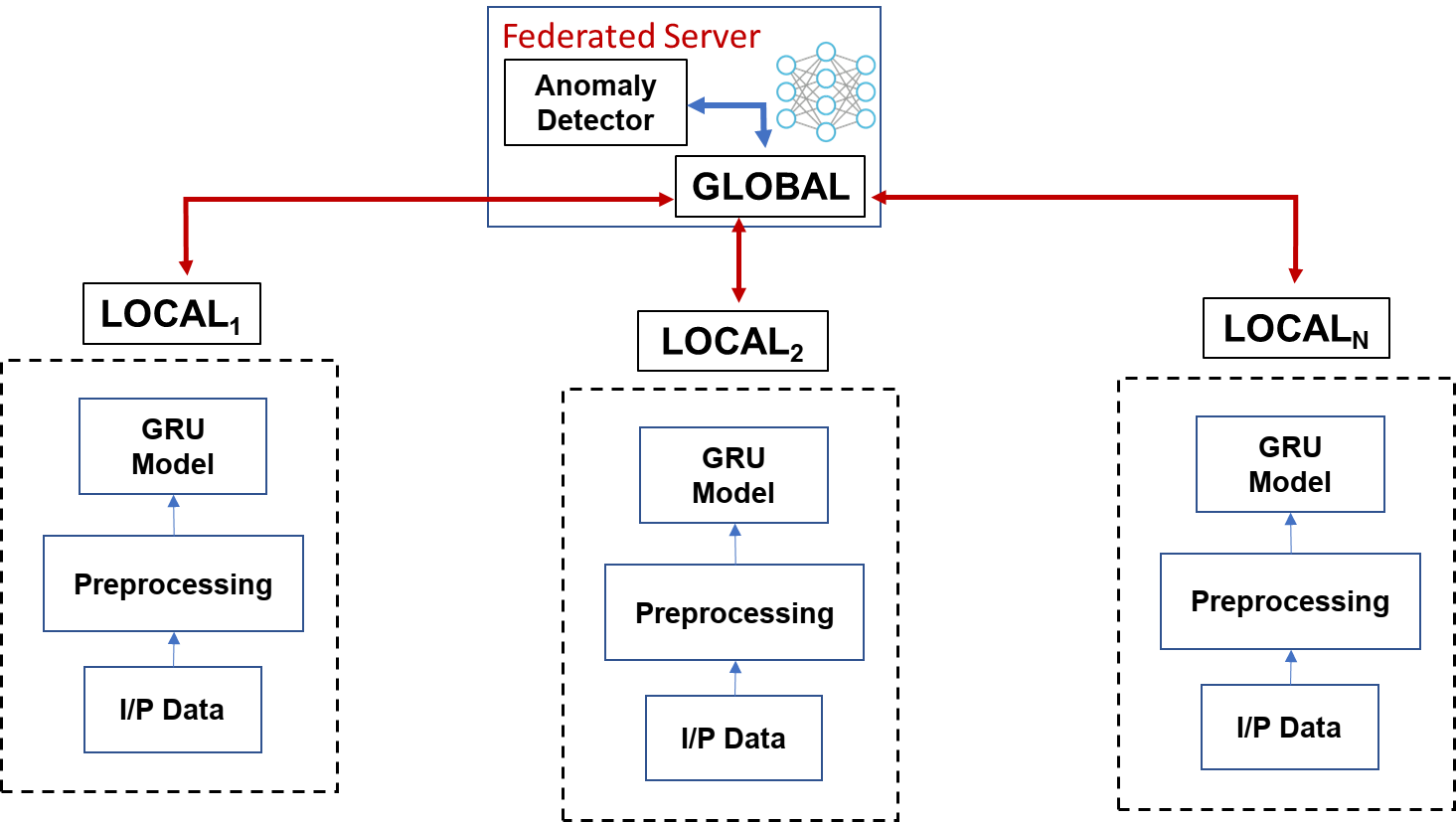}
	\caption{Structure of the FL model}
	\label{fig:Fmodel}
\end{figure}

In our framework, a GRU model (discussed in Section \ref{Sec:GRU}) is learned at each local and global node using the traffic collected, as shown in Fig. \ref{fig:Fmodel}. Each local node is able to perform delay prediction based on the data collected locally. We leverage federated learning at the global node to implement distributed learning from multiple local nodes. It is a privacy-preserving approach where local nodes share model updates with centralized global nodes, rather than collecting data. The IoV network is large with a huge number of edge nodes and contribution from each edge node is imbalanced. To limit the imbalance, we apply clustering to group edge nodes into a set of $E$ number of edge nodes. We use a cluster of six nodes where there are five local nodes and one global server. The choice of global server in each cluster is random and doesn't affect the performance of the network. The idea is to have a single aggregator at the edge level to increase the stronger participation of more vehicles from a distributed IoV network. The aggregation at a global node uses the classical federated averaging function \cite{FL}. The parameters of the global model are updated by local model parameters and calculated as,
\begin{equation}
g=\frac{\sum_{i=1}^{l}k_iw_i}{K}
\end{equation} 
where, $w_i$ is the $i^{th}$ local model parameters, $k_i$ is the respective weight which we use as 1 for all, and $K$ is the number of local nodes.

\subsection{Training Process and Detection}
\begin{figure}[htbp]
	\centering
	\includegraphics[width=3in, height=0.9in]{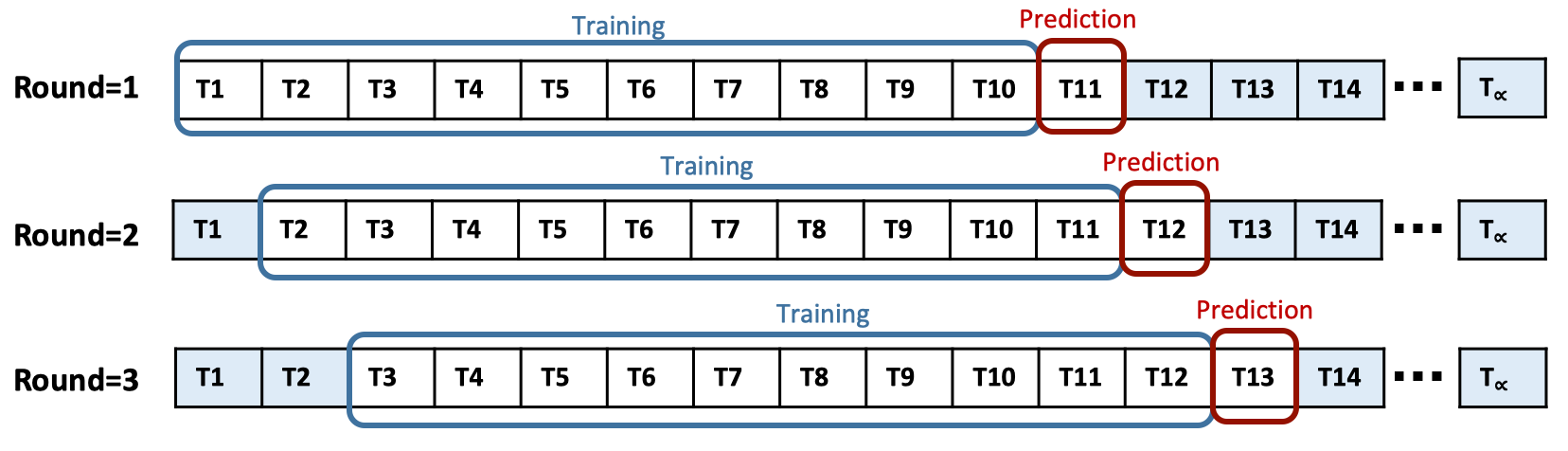}
	\caption{Window-based continual learning}
	\label{fig:trainingRounds}
\end{figure}
The federated training and anomaly detection process are briefly discussed in this section. The training is based on a window-based learning method, as shown in Fig. \ref{fig:trainingRounds}. Each window corresponds to a batch of input values received in multiple time slots. The continual learning works in multiple rounds. In each round, the window slides by one slot. For a given batch size, a set of predictions will be calculated which will help to detect anomalies. In this paper, we use batch size as 200 which predicts a set of 20 (i.e. 10\% of 200) next values with good accuracy. In each round, we roll the window to the right by one slot, update the batch of data by including new set of input values received during the slot, and make a new set of predictions. We illustrate the window-based approach for three rounds in Fig. \ref{fig:trainingRounds}. This process runs in a life-long learning (or continual learning) manner intending to extend the acquired stream of knowledge and use it for future misbehavior detection. \par 
Once we have a set of predictions, the anomaly detection process becomes simple. In the event of an anomaly, the delay observed from vehicles largely varies from the predicted delay. The presence of an anomaly is calculated as,
\begin{equation}
Anomaly=
\begin{cases}
1 & |d_{pred}-d_{rcv}| > D_{th}\\
0 & \text{else}
\end{cases} 
\label{eq:yt}
\end{equation}
where, $d_{pred}$ is the prediction delay, $d_{rcv}$ is the delay received from vehicles, and $D_{th}$ is the delay threshold. The value of $anomaly=1$ indicates the presence of an attack. In this work, we use a delay threshold of 10. The choice of threshold value is important and must work well for attack and no-attack scenarios. If we choose a large threshold, it will be harder for the model to detect the attacks. On the other hand, if we choose a small threshold, multiple false alarms can be triggered. Hence, based on the maximum value of training MSE we get (i.e. $7$ as shown later in results section), we found the threshold of 10 as most appropriate. 
\par 
\begin{algorithm}[htbp]
	\DontPrintSemicolon
	\KwInput{Initialize the global model $\mathcal{M}$ and its parameter $w_g$}
	\For{R=1,2,3,....}
	{	
		Distribute $\mathcal{M}$ and $w_g$ to local nodes \\
		Sample a batch size of $\mathcal{B}$ and preprocess\\
		\For{each client l $\epsilon$ L}
		{	
			$w_l$ $\leftarrow$ \textit{LocalUpdate()} \\
		}
		$w_g=\frac{\sum_{i=1}^{l}w_i}{L}$ \\
	}
	Update global model $\mathcal{M}$ with $w_g$ \\
	Calculate new set of predictions $d'_V$ from $\mathcal{M}$ \\
	Obtain observed delay $d_V$ by vehicles \\
	\If{|$d'_V - d_V$|>$threshold$}{
		$V$ is under attack
	}
	\caption{Global Aggregation and Attack Detection}
	\label{Alg:global}
\end{algorithm}

\begin{algorithm}[htbp]
	\DontPrintSemicolon
	\KwInput{Initialize local model $\mathcal{L}$ = global model $\mathcal{M}$}
	\KwInput{Initialize local model parameter $w_l$}	
	\KwInput{Collect $w_g$ from global}	
	\For{epoch=1,2,3,....200}
	{	
		Sample a batch size of $\mathcal{B}$ and preprocess\\
		$w_l$ $\leftarrow$ Train $\mathcal{L}$ using $w_g$ \\
	}
	return $w_l$ to global server
	\caption{\textit{LocalUpdate()}}
	\label{Alg:local}
\end{algorithm}
We present the global aggregation and attack detection in Algorithm \ref{Alg:global}, and local update in Algorithm \ref{Alg:local}. In Algorithm \ref{Alg:global}, we present the global model aggregation and attack detection function. First, the global model initializes the neural network and its parameters. Line 2 indicates the distribution of the global model to all available local nodes. In line 3, a batch size of $\mathcal{B}$ is selected, and preprocessing is applied as explained in Section \ref{Sec:preprocess}. Lines 4-6 indicate training of local nodes, sending the updated parameters to global node, and a federated average of the global node. In Line 7, the global model updates its parameters using the federated average function. Based on the new and updated model, a set of delay predictions will be calculated in line 8. Finally, the observed delay and predicted delay will be compared to find anomalies in lines 9-11. Algorithm \ref{Alg:local} presents the update of the local node. The local model initializes itself with the received global model and its parameters. Lines 1-3 indicate the training of the local node for 200 epochs with preprocessed data. In line 4, the local model returns the newly-trained local parameters to the global server.
\section{Performance Evaluation}
\label{Sec:performance-evaluation}
In this section, we present performance evaluation results obtained from the extensive simulation of the proposed GFCL anomaly detection framework over an IoV network. We start by describing the experimental setup and performance metrics used to evaluate network performance. Then, we discuss different performance results. 
\subsection{Experimental Setup}
\label{Sec:experimentalsetup}
In this experiment, we use MATLAB software to perform GFCL-based anomaly detection. The GRU architecture of local and global nodes contains an input layer, 3 GRU layers with 64-128-256 hidden unit, a fully connected layer, and a regression layer. The number of hidden components is chosen after performing multiple trial-and-error-based experimentations. We observe in our trials that smaller-to-bigger hidden unit size performs better compared to the bigger-to-smaller hidden unit size in GRU layers. The model is trained on Intel Corei5 2GHz and 8GB RAM CPU using ADAM optimizer. The batch size of 200 is used during every training iteration. The choice of batch size is based on experimental results which are discussed in Section \ref{Sec:batchsize}. The learning rate of 0.01 is used along with a learning rate drop factor of 0.2 after every 125 epochs. The maximum number of the epoch is set to 200. We use a gradient threshold of 1 to prevent the gradients from exploding. The training of the network is continual but in these results, we train our model 10 times (i.e. 10 rounds) with a batch size of 200 during each round of learning. \par 
The vehicle trajectories used in the work are from a real-world vehicle mobility dataset, provided by crawdad \cite{sanfrancisco}. The data is generated from a maximum of $500$ taxis traveling the city of San Francisco. The choice of the dataset is significant, as it is an urban environment with high traffic densities. From the big city area given, we extract an area of $10$x$10$ $km^2$ for use in our experiments. Each taxi is equipped with a GPS sensor and uploads its geo-coordinates record in real-time to form the vehicle trajectories.
\subsection{Performance Metrics}
To verify the performance of our proposed GFCL mechanism, we use the following metrics.
\begin{itemize}
	\item \textit{\textbf{Training Loss:}} It is the indication of error on the training set of data. In our model, we use mean square error (MSE) as our training loss function.  
	\begin{equation}
	\centering
	\mathbb{L}_{train}= \sum_{v} loss
	\end{equation}
	where, $loss$ is calculated using Eq. \ref{Eq:loss}.
	\item \textit{\textbf{MDD - Mean Delay Difference (ms):}} It is the absolute delay difference between the actual delay values and predicted delay values.
	\begin{equation}
	\centering
	MDD\,(ms) = |d_{pred}-d_{rcv}|
	\end{equation}
	where, $d_{pred}$ is the prediction delay and $d_{rcv}$ is the delay received from vehicles.
	\item \textit{\textbf{ACC - Accuracy:}} It indicates the fraction of correct predictions for the test data.
	\begin{equation}
	\centering
	\mathcal{A}CC = \frac{TP+TN}{TP+TN+FP+FN}
	\end{equation}
	where, $TP$, $TN$, $FP$, and $FN$ is true positive, true negative, false positive and false negative, respectively.
	\item \textit{\textbf{DR - Detection Rate:}} It indicates the fraction of correctly detected anomalies among actual attack samples.
	\begin{equation}
	\centering
	\mathcal{DR}= \frac{TP}{TP+FN}
	\end{equation}
	\item \textit{\textbf{FPR - False Positive Rate:}} It indicates how many false anomalies are detected among all no-attack samples. 
	\begin{equation}
	\centering
	\mathcal{FPR}= \frac{FP}{FP+TN}
	\end{equation}
	\item \textit{\textbf{FNR - False Negative Rate:}} It is also known as miss rate and indicates the number of anomalies missed to detect among all anomalies. 
	\begin{equation}
	\centering
	\mathcal{FNR} = \frac{FN}{FN+TP}
	\end{equation}
\end{itemize}

\subsection{Results}
\subsubsection{Performance of Proposed Framework}
\begin{figure}[htbp]
	\centering
	\includegraphics[width=3in, height=2.3in]{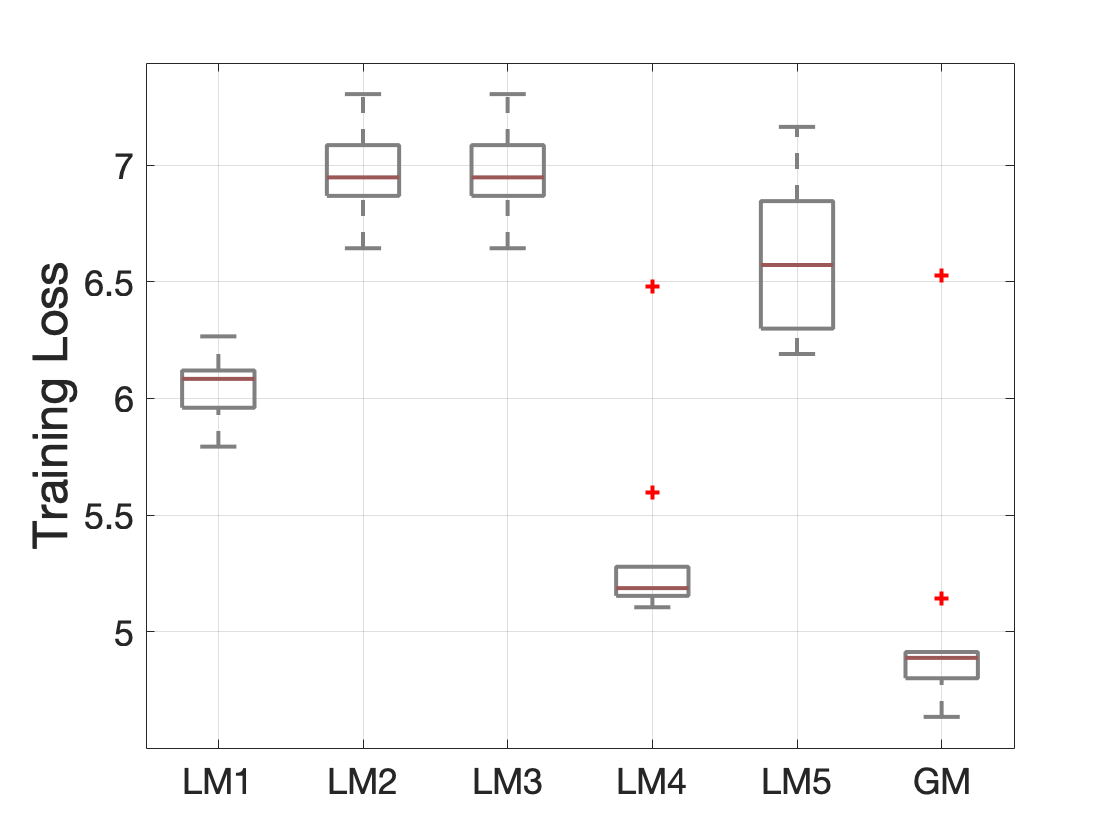}
	\caption{Training Loss}
	\label{fig:trainingMSE}
\end{figure}

\begin{figure}[htbp]
	\centering
	\includegraphics[width=3in, height=2.3in]{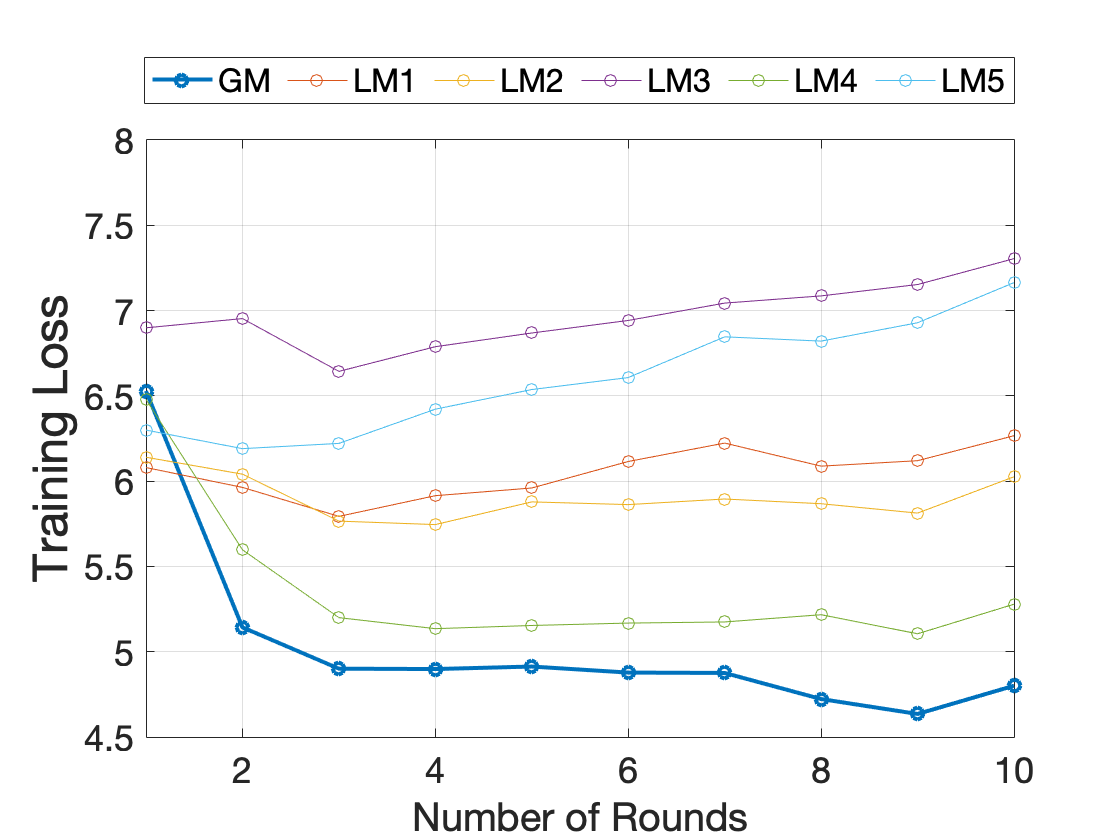}
	\caption{Training Loss vs Number of Rounds}
	\label{fig:trainingMSEvsRounds}
\end{figure}

In this section, we briefly discuss the performance of our proposed GFCL framework using different evaluation metrics. In the first place, Fig. \ref{fig:trainingMSE} and Fig. \ref{fig:trainingMSEvsRounds} illustrate average training loss for different neural network models and the training loss against the number of rounds, respectively. Here, LM and GM denote the local model and the global model, respectively. Fig. \ref{fig:trainingMSE} depicts the average training loss for LMs and GM. The average performance of GM is best compared to all other models. In most of the ML types, the amount of data used to learn has a significant impact on model performance. LMs have limited data in their scope, therefore the training results by local models alone are poorer when compared to the global model. Due to small data, the underfitting problem occurs. On the contrary, in GM, the aggregation of data at a global server maximizes the knowledge and the model performs significantly better than the performance of LM alone. This justifies the use of federated learning along with GRU in our work to enhance the effectiveness of the model by data sharing in V2V. \par
Fig. \ref{fig:trainingMSEvsRounds} shows the training loss against the number of rounds. At the first round of training, the volume of data is small and the model performance is less compared to other rounds of training. This is due to the fact that with little information at first, it is hard for a model to understand the vehicle's pattern and IoV environment. It can also be observed at the first round of training that the performance of GM is poor compared to most of LMs due to the sudden reception of a variety of vehicle patterns from different LMs. However, as the number of rounds increases and the vehicle participation increase, then the loss performance gets significantly better as compared to the first round. In addition, after the first round, the performance of GM is always better than all local models because it has more knowledge compared to all other models. \par

\begin{figure}[htbp]
	\centering
	\includegraphics[width=3in, height=2.3in]{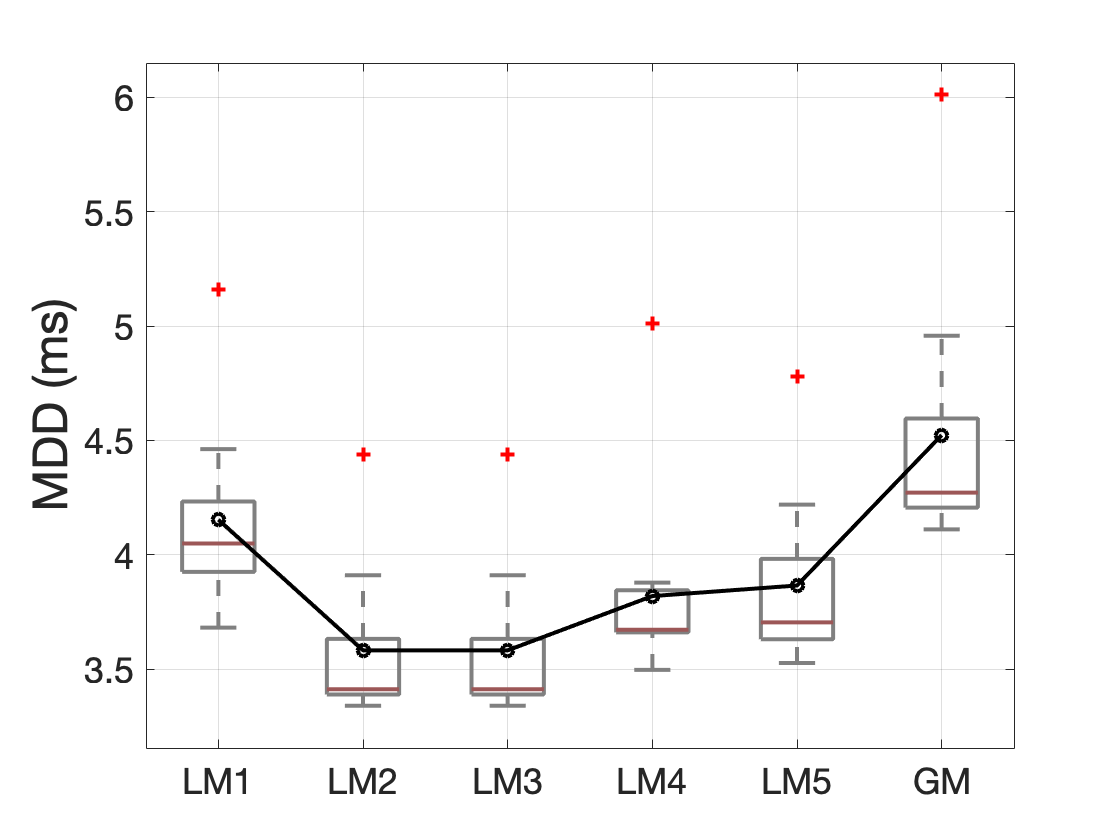}
	\caption{Mean delay difference between predicted value and actual value}
	\label{fig:MDD}
\end{figure}

To analyze the performance of our design in predicting future time slots of a data sequence, we plot mean delay difference (MDD) in Fig. \ref{fig:MDD}. It is the absolute delay difference between actually received delay values and predicted delay values. Notably, the average difference in real and predicted delay is quite low for a given set of predictions, not just a single value of prediction. It is worth noting that the MDD by each node is tightly grouped and it doesn't have many outliers because we chose to perform continual learning to leverage the time-varying property of IoV. It also indicates the good choice of hyper-parameters in our design. \par

\begin{figure}[htbp]
	\centering
	\includegraphics[width=3in, height=2.3in]{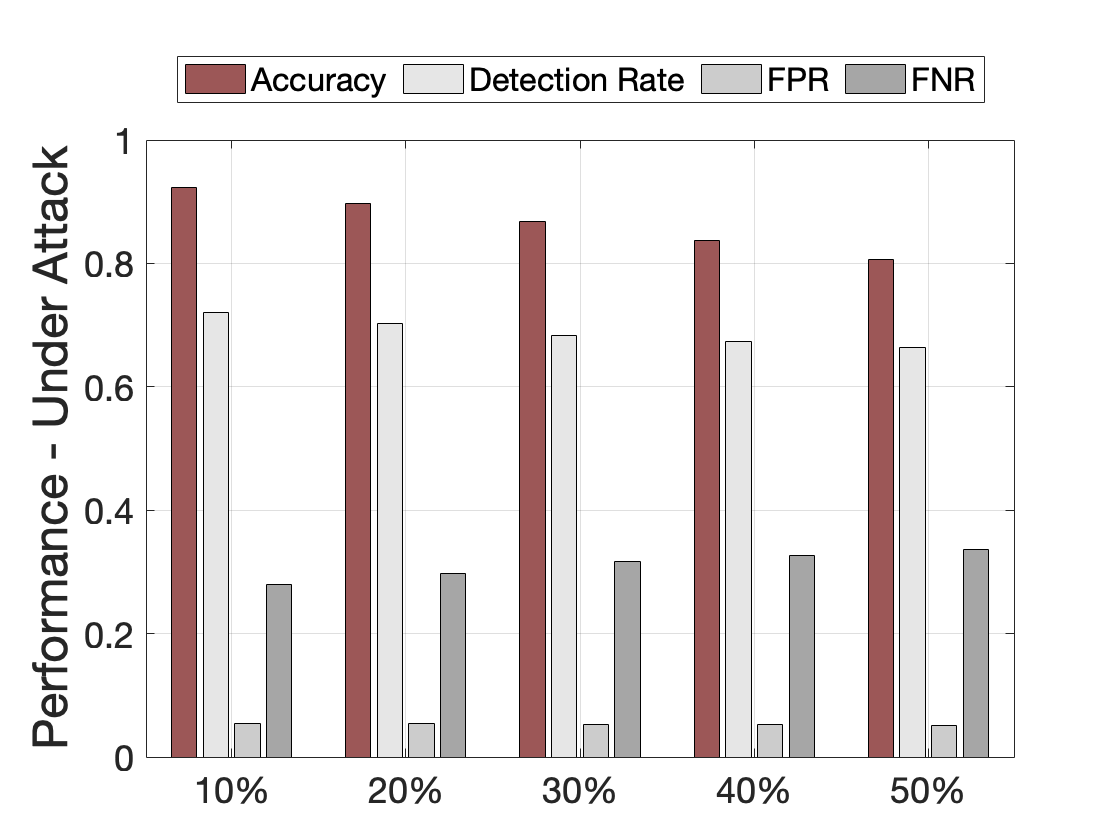}
	\caption{Performance in terms of different metrics vs percentage of attacked vehicles}
	\label{fig:performance}
\end{figure}

Next, we evaluate the performance of the proposed GFCL framework for a real dataset in terms of accuracy, detection rate, FPR, and FNR. These are universally-used metrics to evaluate the efficacy of ML models. The obtained results are listed in Table \ref{tab:performance}. We also plot the accuracy, detection rate, FPR, and FNR performance in Fig. \ref{fig:performance}. We measure the given metrics against different ratios of vehicles under attack. The results show that the proposed GFCL framework achieves good performance in terms of accuracy, detection rate, FPR, and FNR. However, the accuracy has a small reduction when the number of vehicles under attack changes from 10\% to 50\%. While this reduction and the loss in performance is quite small, implying good scalability of our design. \par

\begin{table*}[htbp]
  \centering
  \caption{Framework performance in terms of accuracy (ACC), detection rate (DR), FPR and FNR for vehicle mobilities in the city of San Francisco}
    \begin{tabular}{p{0.4cm}p{0.4cm}p{0.4cm}p{0.4cm}p{0.4cm}|p{0.4cm}p{0.4cm}p{0.4cm}p{0.4cm}|p{0.4cm}p{0.4cm}p{0.4cm}p{0.4cm}|p{0.4cm}p{0.4cm}p{0.4cm}p{0.4cm}|p{0.4cm}p{0.4cm}p{0.4cm}p{0.4cm}}
    \toprule
    \multirow{3}[6]{*}{\textbf{}} & \multicolumn{20}{c}{\textbf{ Percentage of vehicles under attack}} \\
\cmidrule{2-21}    \multicolumn{1}{l}{} & \multicolumn{4}{c}{\textbf{10\%}} & \multicolumn{4}{c}{\textbf{20\%}} & \multicolumn{4}{c}{\textbf{30\%}} & \multicolumn{4}{c}{\textbf{40\%}} & \multicolumn{4}{c}{\textbf{50\%}} \\
\cmidrule{2-21}    \multicolumn{1}{l}{} & \multicolumn{1}{p{0.4cm}}{\textbf{ACC}} & \multicolumn{1}{p{0.4cm}}{\textbf{DR}} & \multicolumn{1}{p{0.4cm}}{\textbf{FPR}} & \multicolumn{1}{p{0.4cm}}{\textbf{FNR}} & \multicolumn{1}{p{0.4cm}}{\textbf{ACC}} & \multicolumn{1}{p{0.4cm}}{\textbf{DR}} & \multicolumn{1}{p{0.4cm}}{\textbf{FPR}} & \multicolumn{1}{p{0.4cm}}{\textbf{FNR}} & \multicolumn{1}{p{0.4cm}}{\textbf{ACC}} & \multicolumn{1}{p{0.4cm}}{\textbf{DR}} & \multicolumn{1}{p{0.4cm}}{\textbf{FPR}} & \multicolumn{1}{p{0.4cm}}{\textbf{FNR}} & \multicolumn{1}{p{0.4cm}}{\textbf{ACC}} & \multicolumn{1}{p{0.4cm}}{\textbf{DR}} & \multicolumn{1}{p{0.4cm}}{\textbf{FPR}} & \multicolumn{1}{p{0.4cm}}{\textbf{FNR}} & \multicolumn{1}{p{0.4cm}}{\textbf{ACC}} & \multicolumn{1}{p{0.4cm}}{\textbf{DR}} & \multicolumn{1}{p{0.4cm}}{\textbf{FPR}} & \textbf{FNR} \\
    \midrule
    \textbf{\textit{R1}} & 0.92  & 0.66  & 0.051 & 0.34  & 0.892 & 0.68  & 0.055 & 0.32  & 0.87  & 0.686 & 0.051 & 0.31 & 0.838 & 0.67  & 0.05  & 0.33  & 0.806 & 0.66  & 0.048 & 0.34 \\
    \textbf{\textit{R2}} & 0.926 & 0.68  & 0.046 & 0.32  & 0.908 & 0.71  & 0.042 & 0.29  & 0.89  & 0.713 & 0.04  & 0.28 & 0.87  & 0.73  & 0.036 & 0.27  & 0.84  & 0.708 & 0.028 & 0.29 \\
    \textbf{\textit{R3}} & 0.926 & 0.72  & 0.051 & 0.28  & 0.89  & 0.68  & 0.057 & 0.32  & 0.85  & 0.653 & 0.06  & 0.34 & 0.822 & 0.65  & 0.063 & 0.35  & 0.784 & 0.632 & 0.064 & 0.36 \\
    \textbf{\textit{R4}} & 0.942 & 0.74  & 0.035 & 0.26  & 0.914 & 0.71  & 0.035 & 0.29  & 0.87  & 0.64  & 0.028 & 0.36  & 0.832 & 0.62  & 0.026 & 0.38  & 0.794 & 0.62  & 0.032 & 0.38 \\
    \textbf{\textit{R5}} & 0.934 & 0.74  & 0.044 & 0.26  & 0.902 & 0.69  & 0.045 & 0.31  & 0.87  & 0.673 & 0.048 & 0.32 & 0.838 & 0.665 & 0.046 & 0.33 & 0.796 & 0.644 & 0.052 & 0.35 \\
    \textbf{\textit{R6}} & 0.936 & 0.62  & 0.028 & 0.38  & 0.92  & 0.71  & 0.027 & 0.29  & 0.89  & 0.693 & 0.028 & 0.30 & 0.858 & 0.695 & 0.033 & 0.30 & 0.84  & 0.704 & 0.024 & 0.29 \\
    \textbf{\textit{R7}} & 0.94  & 0.88  & 0.053 & 0.12  & 0.90   & 0.72  & 0.055 & 0.28  & 0.86  & 0.673 & 0.054 & 0.32 & 0.822 & 0.645 & 0.06  & 0.35 & 0.788 & 0.636 & 0.06  & 0.36 \\
    \textbf{\textit{R8}} & 0.93  & 0.74  & 0.049 & 0.26  & 0.904 & 0.71  & 0.047 & 0.29  & 0.87  & 0.693 & 0.051 & 0.30 & 0.842 & 0.68  & 0.05  & 0.32  & 0.822 & 0.692 & 0.048 & 0.30 \\
    \textbf{\textit{R9}} & 0.868 & 0.70   & 0.113 & 0.30   & 0.85  & 0.68  & 0.107 & 0.32  & 0.82  & 0.66  & 0.114 & 0.34  & 0.798 & 0.665 & 0.113 & 0.33 & 0.782 & 0.664 & 0.10   & 0.33 \\
    \textbf{\textit{R10}} & 0.90   & 0.68  & 0.075 & 0.32  & 0.896 & 0.73  & 0.062 & 0.27  & 0.87  & 0.706 & 0.057 & 0.29 & 0.842 & 0.685 & 0.053 & 0.31 & 0.81  & 0.68  & 0.06  & 0.32 \\
    \bottomrule
    \end{tabular}%
  \label{tab:performance}%
\end{table*}%

\subsubsection{Impact of Batch Size}
\begin{figure}[htbp]
	\centering
	\includegraphics[width=3in, height=2.3in]{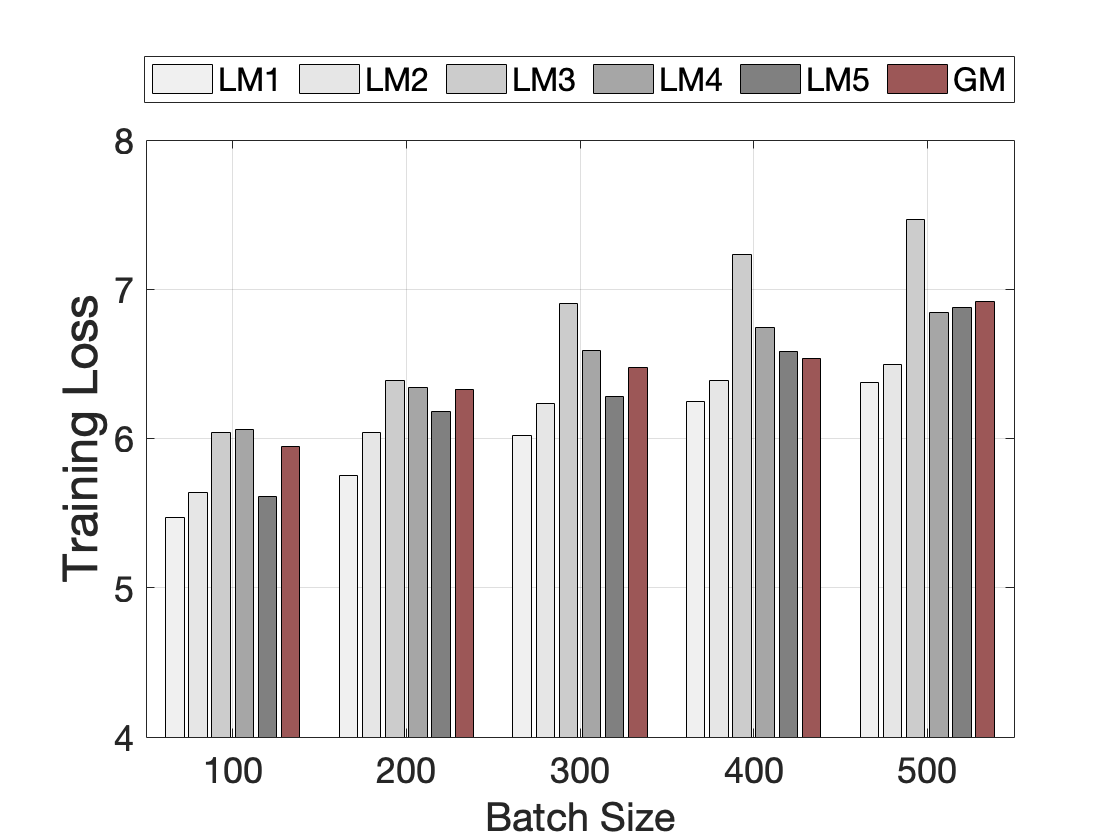}
	\caption{Training Loss vs Batch Size}
	\label{fig:TLBS}
\end{figure}

There are multiple important hyper-parameters in the design of a neural network, among which the batch size is important and crucial when we are dealing with GRU where prediction of future time slots is performed based on input values. For a given design, we optimize our network such that for each input sequence, the GRU model must provide a good prediction accuracy for an output prediction sequence with a size equal to 10\% of input. As an example, if the input sequence is of size 100 then the model must maintain good prediction accuracy for output of size 10 (i.e. 10\% of 100). Fig. \ref{fig:TLBS} shows the training loss for local and global models against different batch sizes. We study five different batch sizes and compare the performance among them. According to our experimental results, with an increase in batch size, training loss increases. This is due to the fact that the size of the prediction set also increases and impacts the average performance of the network. To maintain a good balance of lower training loss and a bigger set of predictions, we choose to use a batch size of 200 in our experiments.
\label{Sec:batchsize}

\subsubsection{Impact of Different Datasets}
In addition to the San Francisco vehicle mobility dataset, we evaluated our model with two other real-world vehicle mobility datasets where vehicles are traveling in the city of Rome \cite{rome} and Beijing \cite{beijing1,beijing2}, respectively. This is to study and verify the effectiveness of our model for varying vehicle densities and different vehicular environments. From the big city area of Rome and Beijing given, we extract an area of $10$x$10$ $km^2$ from each dataset for use in our experiments. The data is generated from a maximum of $174$ taxis and $205$ taxis traveling the city of Rome and Beijing, respectively. The number of active taxis at a given time slot also varies. The choice of the datasets is significant, as both are urban environments with different traffic densities compared to San Francisco's mobility traces. \par
\begin{figure}[htbp]
	\centering
	\begin{subfigure}{.24\textwidth}
		\centering
		% include first image
		\includegraphics[width=1.9in,height=1.45in]{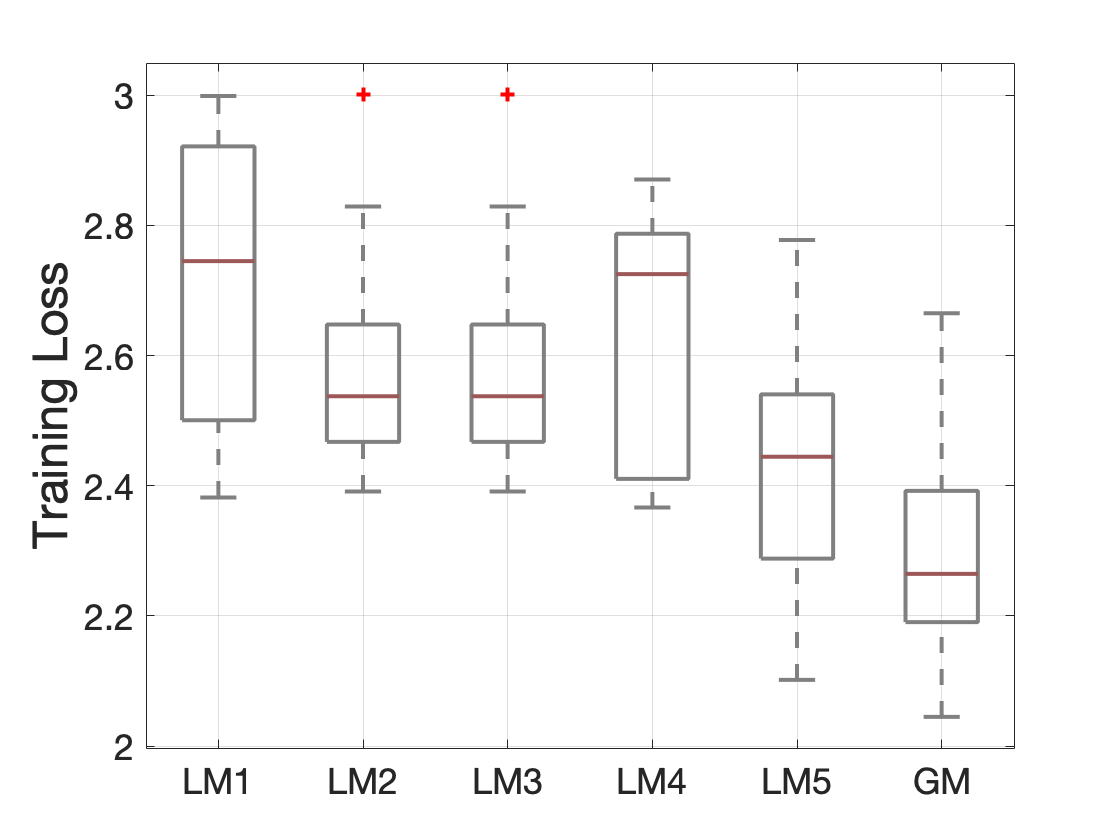}  
		\caption{Rome}
		\label{fig:TLrome}
	\end{subfigure}
	\begin{subfigure}{.24\textwidth}
		\centering
		% include first image
		\includegraphics[width=1.9in,height=1.45in]{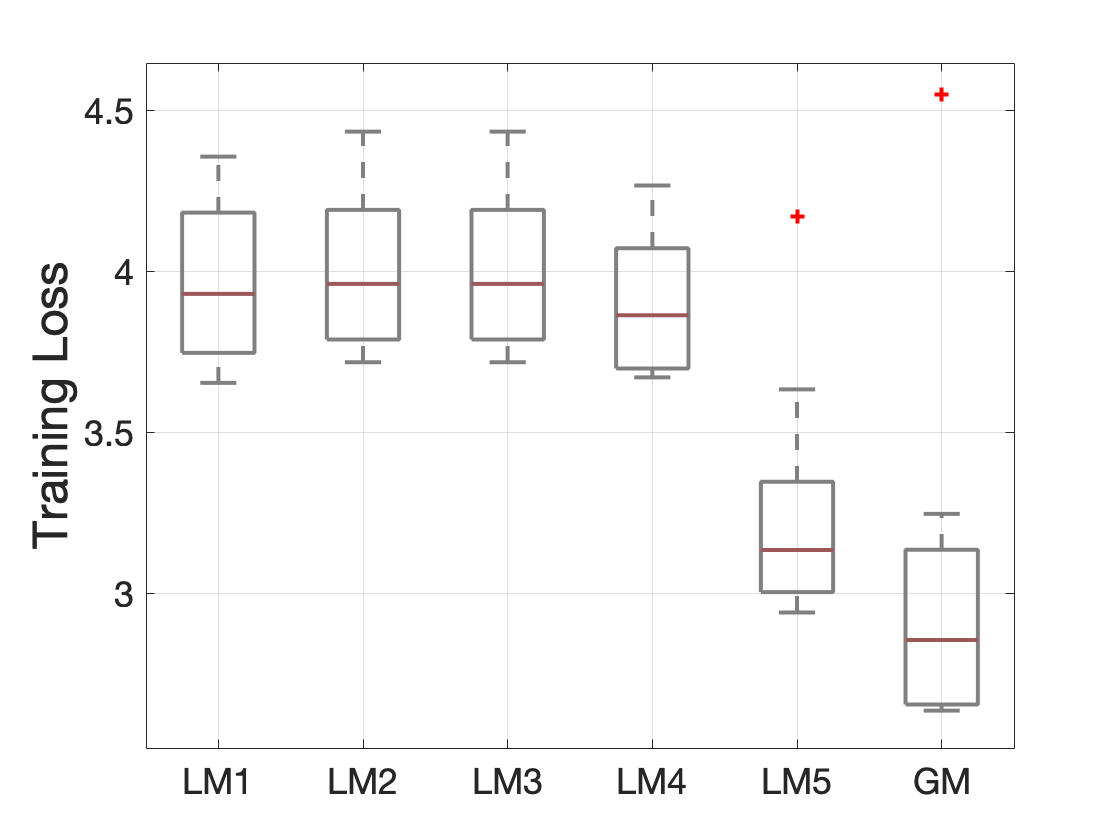}  
		\caption{Beijing}
		\label{fig:TLbeijing}
	\end{subfigure}
	\caption{Training Loss}
	\label{fig:TLdatasets}
\end{figure}

 
Furthermore, we compare the performance of our model for different datasets in terms of training loss, accuracy, detection rate, FPR, and FNR. Fig. \ref{fig:TLrome} and Fig. \ref{fig:TLbeijing} depict the training loss for the vehicle mobilities of Rome and Beijing city, respectively. Interestingly, with the same design parameters, our proposed anomaly detection framework shows quite low training loss even though the vehicle density and city environment are changed. The trend of GM and LM is also obvious where the average loss for GM is better than any of the local models. \par

We also compare the accuracy, detection rate, FPR, and FNR of our framework for Rome and Beijing city environment in Table \ref{tab:performanceRome} and Table \ref{tab:performanceBeijing}, respectively We can observe that our proposed GFCL framework, which is now trained on two different city environments, achieves good performance against different ratios of vehicles under attack. The performance of FPR in the Rome city environment is significant where no false positives are reported. The accuracy and detection rate achieved is also noteworthy for both cities, where even with 50\% of vehicles under attack, the average accuracy doesn't fall below 86\% and 80\% for Rome and Beijing city, respectively, and the average detection rate doesn't fall below 71\% and 69\% Rome and Beijing city, respectively. The good performance in terms of the above metrics for different city environments and dynamic traffic densities proves high scalability of our proposed solution framework.

\begin{table*}[htbp]
	\centering
	\caption{Framework performance in terms of accuracy (ACC), detection rate (DR), FPR and FNR for vehicle mobilities in the Rome city}
	\begin{tabular}{p{0.4cm}p{0.4cm}p{0.4cm}cp{0.4cm}|p{0.4cm}p{0.4cm}cp{0.4cm}|p{0.4cm}p{0.4cm}cp{0.4cm}|p{0.4cm}p{0.4cm}cp{0.4cm}|p{0.4cm}p{0.4cm}cp{0.4cm}}
		\toprule
		\multirow{3}[6]{*}{\textbf{}} & \multicolumn{20}{c}{\textbf{ Percentage of vehicles under attack}} \\
		\cmidrule{2-21}    \multicolumn{1}{l}{} & \multicolumn{4}{c}{\textbf{10\%}} & \multicolumn{4}{c}{\textbf{20\%}} & \multicolumn{4}{c}{\textbf{30\%}} & \multicolumn{4}{c}{\textbf{40\%}} & \multicolumn{4}{c}{\textbf{50\%}} \\
		\cmidrule{2-21}    \multicolumn{1}{l}{} & \multicolumn{1}{p{0.4cm}}{\textbf{ACC}} & \multicolumn{1}{p{0.4cm}}{\textbf{DR}} & \multicolumn{1}{p{0.4cm}}{\textbf{FPR}} & \multicolumn{1}{p{0.4cm}}{\textbf{FNR}} & \multicolumn{1}{p{0.4cm}}{\textbf{ACC}} & \multicolumn{1}{p{0.4cm}}{\textbf{DR}} & \multicolumn{1}{p{0.4cm}}{\textbf{FPR}} & \multicolumn{1}{p{0.4cm}}{\textbf{FNR}} & \multicolumn{1}{p{0.4cm}}{\textbf{ACC}} & \multicolumn{1}{p{0.4cm}}{\textbf{DR}} & \multicolumn{1}{p{0.4cm}}{\textbf{FPR}} & \multicolumn{1}{p{0.4cm}}{\textbf{FNR}} & \multicolumn{1}{p{0.4cm}}{\textbf{ACC}} & \multicolumn{1}{p{0.4cm}}{\textbf{DR}} & \multicolumn{1}{p{0.4cm}}{\textbf{FPR}} & \multicolumn{1}{p{0.4cm}}{\textbf{FNR}} & \multicolumn{1}{p{0.4cm}}{\textbf{ACC}} & \multicolumn{1}{p{0.4cm}}{\textbf{DR}} & \multicolumn{1}{p{0.4cm}}{\textbf{FPR}} & \textbf{FNR} \\
		\midrule
		\textbf{\textit{R1}} & 0.953 & 0.533 & 0     & 0.46 & 0.933 & 0.667 & 0     & 0.33  & 0.92  & 0.733 & 0     & 0.26 & 0.887 & 0.717 & 0     & 0.28  & 0.84  & 0.68  & 0     & 0.32 \\
		\textbf{\textit{R2}} & 0.96  & 0.6   & 0     & 0.4   & 0.953 & 0.767 & 0     & 0.23 & 0.933 & 0.778 & 0     & 0.22  & 0.907 & 0.767 & 0     & 0.23 & 0.867 & 0.733 & 0     & 0.26 \\
		\textbf{\textit{R3}} & 0.967 & 0.667 & 0     & 0.33 & 0.947 & 0.733 & 0     & 0.26 & 0.933 & 0.778 & 0     & 0.22  & 0.893 & 0.73  & 0     & 0.26 & 0.867 & 0.733 & 0     & 0.26 \\
		\textbf{\textit{R4}} & 0.967 & 0.667 & 0     & 0.33 & 0.94  & 0.7   & 0     & 0.3   & 0.933 & 0.778 & 0     & 0.22  & 0.907 & 0.767 & 0     & 0.23 & 0.867 & 0.733 & 0     & 0.26 \\
		\textbf{\textit{R5}} & 0.98  & 0.8   & 0     & 0.2   & 0.94  & 0.7   & 0     & 0.3   & 0.933 & 0.778 & 0     & 0.22  & 0.913 & 0.783 & 0     & 0.22 & 0.867 & 0.733 & 0     & 0.26 \\
		\textbf{\textit{R6}} & 0.973 & 0.733 & 0     & 0.26 & 0.94  & 0.7   & 0     & 0.3   & 0.913 & 0.711 & 0     & 0.29 & 0.887 & 0.717 & 0     & 0.28 & 0.853 & 0.707 & 0     & 0.29 \\
		\textbf{\textit{R7}} & 0.967 & 0.667 & 0     & 0.33 & 0.927 & 0.633 & 0     & 0.37 & 0.913 & 0.711 & 0     & 0.29 & 0.88  & 0.7   & 0     & 0.3   & 0.84  & 0.68  & 0     & 0.32 \\
		\textbf{\textit{R8}} & 0.987 & 0.867 & 0     & 0.13 & 0.94  & 0.7   & 0     & 0.3   & 0.907 & 0.689 & 0     & 0.31 & 0.873 & 0.683 & 0     & 0.32 & 0.84  & 0.68  & 0     & 0.32 \\
		\textbf{\textit{R9}} & 0.987 & 0.867 & 0     & 0.13 & 0.953 & 0.767 & 0     & 0.23 & 0.92  & 0.733 & 0     & 0.26 & 0.9   & 0.75  & 0     & 0.25  & 0.853 & 0.707 & 0     & 0.29 \\
		\textbf{\textit{R10}} & 0.96  & 0.6   & 0     & 0.4   & 0.94  & 0.7   & 0     & 0.3   & 0.893 & 0.644 & 0     & 0.35 & 0.887 & 0.717 & 0     & 0.28 & 0.847 & 0.693 & 0     & 0.31 \\
		\bottomrule
	\end{tabular}%
	\label{tab:performanceRome}%
\end{table*}%

\begin{table*}[htbp]
	\centering
	\caption{Framework performance in terms of accuracy (ACC), detection rate (DR), FPR and FNR for vehicle mobilities in the Beijing city}
	\begin{tabular}{p{0.4cm}p{0.4cm}p{0.4cm}p{0.4cm}p{0.4cm}|p{0.4cm}p{0.4cm}p{0.4cm}p{0.4cm}|p{0.4cm}p{0.4cm}p{0.4cm}p{0.4cm}|p{0.4cm}p{0.4cm}p{0.4cm}p{0.4cm}|p{0.4cm}p{0.4cm}p{0.4cm}p{0.4cm}}
		\toprule
		\multirow{3}[6]{*}{\textbf{}} & \multicolumn{20}{c}{\textbf{ Percentage of vehicles under attack}} \\
		\cmidrule{2-21}    \multicolumn{1}{l}{} & \multicolumn{4}{c}{\textbf{10\%}} & \multicolumn{4}{c}{\textbf{20\%}} & \multicolumn{4}{c}{\textbf{30\%}} & \multicolumn{4}{c}{\textbf{40\%}} & \multicolumn{4}{c}{\textbf{50\%}} \\
		\cmidrule{2-21}    \multicolumn{1}{l}{} & \multicolumn{1}{p{0.4cm}}{\textbf{ACC}} & \multicolumn{1}{p{0.4cm}}{\textbf{DR}} & \multicolumn{1}{p{0.4cm}}{\textbf{FPR}} & \multicolumn{1}{p{0.4cm}}{\textbf{FNR}} & \multicolumn{1}{p{0.4cm}}{\textbf{ACC}} & \multicolumn{1}{p{0.4cm}}{\textbf{DR}} & \multicolumn{1}{p{0.4cm}}{\textbf{FPR}} & \multicolumn{1}{p{0.4cm}}{\textbf{FNR}} & \multicolumn{1}{p{0.4cm}}{\textbf{ACC}} & \multicolumn{1}{p{0.4cm}}{\textbf{DR}} & \multicolumn{1}{p{0.4cm}}{\textbf{FPR}} & \multicolumn{1}{p{0.4cm}}{\textbf{FNR}} & \multicolumn{1}{p{0.4cm}}{\textbf{ACC}} & \multicolumn{1}{p{0.4cm}}{\textbf{DR}} & \multicolumn{1}{p{0.4cm}}{\textbf{FPR}} & \multicolumn{1}{p{0.4cm}}{\textbf{FNR}} & \multicolumn{1}{p{0.4cm}}{\textbf{ACC}} & \multicolumn{1}{p{0.4cm}}{\textbf{DR}} & \multicolumn{1}{p{0.4cm}}{\textbf{FPR}} & \textbf{FNR} \\
		\midrule
		\textbf{\textit{R1}} & 0.92  & 0.76  & 0.054 & 0.23  & 0.91  & 0.76  & 0.055 & 0.24  & 0.87  & 0.68  & 0.04  & 0.32  & 0.85  & 0.69  & 0.049 & 0.3   & 0.81  & 0.69  & 0.059 & 0.31 \\
		\textbf{\textit{R2}} & 0.89  & 0.619 & 0.07  & 0.38  & 0.87  & 0.66  & 0.073 & 0.34  & 0.82  & 0.6   & 0.07  & 0.4   & 0.8   & 0.622 & 0.081 & 0.38  & 0.76  & 0.62  & 0.098 & 0.37 \\
		\textbf{\textit{R3}} & 0.92  & 0.76  & 0.054 & 0.24 & 0.9   & 0.756 & 0.061 & 0.24  & 0.87  & 0.71  & 0.06  & 0.29  & 0.85  & 0.73  & 0.07  & 0.27 & 0.82  & 0.71  & 0.06  & 0.29 \\
		\textbf{\textit{R4}} & 0.92  & 0.857 & 0.065 & 0.14 & 0.9   & 0.8   & 0.073 & 0.19  & 0.87  & 0.74  & 0.08  & 0.26  & 0.85  & 0.768 & 0.08  & 0.23  & 0.83  & 0.76  & 0.098 & 0.24 \\
		\textbf{\textit{R5}} & 0.94  & 0.71  & 0.032 & 0.28  & 0.93  & 0.78  & 0.03  & 0.21  & 0.88  & 0.68  & 0.02  & 0.32  & 0.87  & 0.719 & 0.02  & 0.28  & 0.85  & 0.74  & 0.02  & 0.26 \\
		\textbf{\textit{R6}} & 0.93  & 0.71  & 0.038 & 0.28  & 0.93  & 0.78  & 0.036 & 0.21  & 0.89  & 0.74  & 0.04  & 0.26  & 0.86  & 0.73  & 0.04  & 0.26  & 0.85  & 0.767 & 0.05  & 0.233 \\
		\textbf{\textit{R7}} & 0.81  & 0.71  & 0.16  & 0.28 & 0.82  & 0.8   & 0.176 & 0.19  & 0.8   & 0.72  & 0.16  & 0.27  & 0.81  & 0.756 & 0.15  & 0.24  & 0.8   & 0.747 & 0.15  & 0.252 \\
		\textbf{\textit{R8}} & 0.68  & 0.38  & 0.27  & 0.61  & 0.697 & 0.512 & 0.25  & 0.48  & 0.7   & 0.6   & 0.25  & 0.4   & 0.69  & 0.622 & 0.26  & 0.37  & 0.69  & 0.66  & 0.27  & 0.339 \\
		\textbf{\textit{R9}} & 0.74  & 0.142 & 0.19  & 0.86 & 0.73  & 0.39  & 0.18  & 0.6   & 0.73  & 0.55  & 0.18  & 0.45  & 0.717 & 0.57  & 0.18  & 0.42  & 0.74  & 0.64  & 0.15  & 0.359 \\
		\textbf{\textit{R10}} & 0.73  & 0     & 0.179 & 1     & 0.736 & 0.268 & 0.14  & 0.73  & 0.75  & 0.47  & 0.12  & 0.53  & 0.746 & 0.52  & 0.105 & 0.47  & 0.75  & 0.55  & 0.05  & 0.44 \\
		\bottomrule
	\end{tabular}%
	\label{tab:performanceBeijing}%
\end{table*}%

\section{Conclusion}
\label{Sec:conclusion}
In this paper, we addressed the problem of Sybil-based data poisoning attacks in the context of DRL-based IoV applications. We use the GRU-based regression model to predict the future data sequence to analyze and detect illegitimate behavior from vehicles. Typically, an anomaly detection mechanism on a single node with limited data may suffer from low sensitivity or high false positives when a wide variety of vehicle behavior is captured. Therefore, we proposed to integrate federated learning with GRU to widen the knowledge and enhance the performance of the anomaly detection framework. We evaluated our framework on several real-world vehicle mobility traces. We carried out an extensive set of experiments to demonstrate low training loss and high detection accuracy of illegitimate nodes for different fractions of vehicles under attack. We also verified the scalability of our design with good performance in different city environments.

\balance

% references section
\bibliographystyle{IEEEtran}
\bibliography{IEEEabrv,References}

\end{document}